\documentclass[10pt,twocolumn,letterpaper]{article}

\usepackage{iccv}
\usepackage{times}
\usepackage{epsfig}
\usepackage{graphicx}
\usepackage{amsmath}
\usepackage{amssymb}
\usepackage{caption,subcaption}
\usepackage{pifont}
\usepackage{multirow}
\usepackage{algorithm}
\usepackage{algpseudocode}
\usepackage[pagebackref=true,breaklinks=true,letterpaper=true,colorlinks,bookmarks=false]{hyperref}

\iccvfinalcopy % *** Uncomment this line for the final submission

\def\iccvPaperID{3317} % *** Enter the ICCV Paper ID here

\ificcvfinal\fi

\begin{document}

%%%%%%%%% TITLE
\title{Learnable Boundary Guided Adversarial Training}

\author{
	Jiequan Cui $^{1}$ \quad 
	Shu Liu $^{2}$ \quad
	Liwei Wang $^{1}$ \quad
	Jiaya Jia $^{1,2}$ \\
	$^{1}$The Chinese University of Hong Kong \hspace{1cm} $^{2}$SmartMore \hspace{1cm} 
	\vspace{.7em}\\
	{\tt\small \{jqcui, lwwang, leojia\}@cse.cuhk.edu.hk, liushuhust@gmail.com}
}

\maketitle
\ificcvfinal\thispagestyle{empty}\fi

%%%%%%%%% ABSTRACT
\begin{abstract}
   Previous adversarial training raises model robustness under the compromise of accuracy on natural data. In this paper, we reduce natural accuracy degradation. We use the model logits from one clean model to guide learning of another one robust model, taking into consideration that logits from the well trained clean model embed the most discriminative features of natural data, {\it e.g.}, generalizable classifier boundary. Our solution is to constrain logits from the robust model that takes adversarial examples as input and makes it similar to those from the clean model fed with corresponding natural data. It lets the robust model inherit the classifier boundary of the clean model. Moreover, we observe such boundary guidance can not only preserve high natural accuracy but also benefit model robustness, which gives new insights and facilitates progress for the adversarial community. Finally, extensive experiments on CIFAR-10, CIFAR-100, and Tiny ImageNet testify to the effectiveness of our method. We achieve new state-of-the-art robustness on CIFAR-100 without additional real or synthetic data with auto-attack benchmark \footnote{\url{https://github.com/fra31/auto-attack}}. Our code is available at \url{https://github.com/dvlab-research/LBGAT}.
\end{abstract}

%%%%%%%%% BODY TEXT
\section{Introduction}
Deep neural networks have achieved great success in many tasks, especially with the surge of neural architecture search \cite{DBLP:conf/cvpr/ZophVSL18, DBLP:conf/iclr/LiuSY19, DBLP:conf/cvpr/TanCPVSHL19, DBLP:conf/iccv/CuiCLLSJ19, DBLP:conf/iclr/CaiGWZH20}. However, with the concern of security of deep models, several methods~\cite{DBLP:conf/cvpr/DongLPS0HL18,Xie_2019_CVPR,DBLP:journals/corr/SzegedyZSBEGF13,Shi_2019_CVPR,DBLP:conf/iclr/TramerKPGBM18,DBLP:conf/aaai/ZhengC019,DBLP:conf/iclr/TramerKPGBM18,DBLP:conf/cvpr/HeZRS16,DBLP:conf/cvpr/HuangLMW17,DBLP:journals/corr/SimonyanZ14a} have shown that deep models could be vulnerable to adversarial attack. Data that is intentionally created may easily fool strong classifiers. 

\begin{figure}
	\begin{center}
		\includegraphics[width=0.48\textwidth]{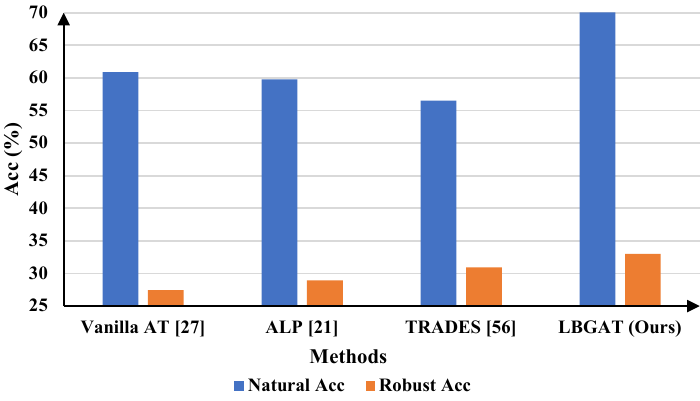}
		\caption{Model robustness on CIFAR-100 evaluated with 20 iterations PGD under white-box attack. ``Natural Acc'' represents classification accuracy on natural (clean) data. ``Robust Acc'' represents classification accuracy on adversarial data.
			Our method (LBGAT+TRADES with $\alpha=0$) improves robustness with the least natural accuracy degradation.}
		\label{fig:comparison_methods}
	\end{center}
	\vspace{-0.3in}
\end{figure}

In response to the vulnerability of deep neural networks, adversarial defense has become an essential topic in computer vision. There are now a sizable body of work exploring different ways to get adversarial settings, including defensive distillation \cite{DBLP:conf/sp/PapernotM0JS16}, feature squeezing \cite{DBLP:conf/ndss/Xu0Q18}, randomization based methods \cite{DBLP:conf/iclr/XieWZRY18,DBLP:conf/iclr/DhillonALBKKA18} and augmenting the training with adversarial examples~\cite{zhang2019theoretically,DBLP:journals/corr/abs-1803-06373,DBLP:conf/iclr/MadryMSTV18,DBLP:conf/iclr/TramerKPGBM18}, {\it i.e.}, adversarial training. However, training a robust model is still challenging. Recently, adversarial training with PGD attack~\cite{DBLP:conf/iclr/MadryMSTV18} becomes an effective defense strategy. However, when plotting results of recent work~\cite{zhang2019theoretically,DBLP:journals/corr/abs-1803-06373,DBLP:conf/iclr/MadryMSTV18} in Fig. \ref{fig:comparison_methods}, it is still noticeable that higher robustness is often accompanied with more accuracy degradation on natural data classification. 

Different from previous work that mainly pursues various ways to improve robustness, we meanwhile pursue accuracy preservation on natural data.
In this paper, we propose a novel adversarial training scheme, which significantly improves classification accuracy on natural data. It also achieves high robustness under black- and white-box attack. We take advantage of logits from a clean model, which is trained only on natural data, to guide the learning of a robust model. 

\begin{figure*}[t]
	\begin{center}
		\includegraphics[width=1.01\textwidth]{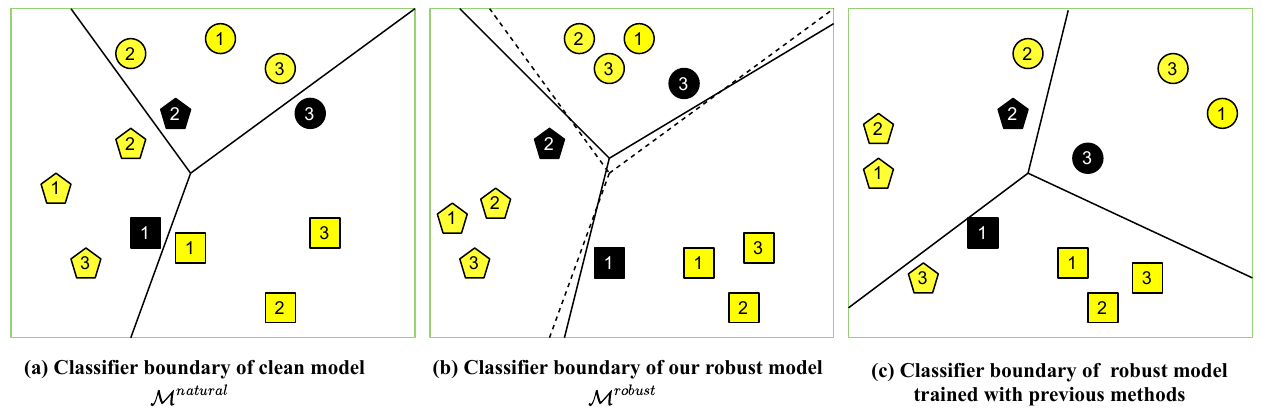}
		\caption{Conceptual illustration of our method vs. previous adversarial training approaches. Solid lines denote real classifier boundary of the trained model, while the dotted line is the classifier boundary of the clean model $\mathcal{M}^{natural}$. Different shapes represent logits of images in various classes. Black color marks adversarial examples.} 
		\label{fig:illustration}
	\end{center}
	\vspace{-0.2in}
\end{figure*}

A conceptual illustration is shown in Fig. \ref{fig:illustration} to explain our motivation. As shown in (a), when only trained on natural (clean) data, the learned model $\mathcal{M}^{natural}$ separates natural data (plotted in yellow) well. But it may fail to classify perturbed data and misclassifies the dark circle into the rectangle category. Previous standard adversarial training methods, {\it e.g.}, Madry et al. \cite{DBLP:conf/iclr/MadryMSTV18}, mainly improve the robustness towards adversarial examples. As shown in Fig. \ref{fig:illustration}(c), adversarial examples (plotted in black) can be mostly correctly classified with this strategy. However, some clean data is wrong. Thus, our motivation is to leverage the clean model $\mathcal{M}^{natural}$ to improve the natural data accuracy of $\mathcal{M}^{robust}$.

In order to seek guidance from clean model $\mathcal{M}^{natural}$, we expect the logit output of adversarial example $x^{adv}$ from $\mathcal{M}^{robust}$ to be similar to logits output of corresponding natural data $x$ that goes through $\mathcal{M}^{natural}$. As plotted in Fig. \ref{fig:illustration}(b), the classifier boundary of our $\mathcal{M}^{robust}$ is constrained by that of the clean model, which helps classify the clean data into correct categories. At the same time, adversarial examples are also correctly labeled, benefiting from the adversarial training scheme.

Instead of constraining $\mathcal{M}^{robust}$ with the classifier boundary from one well trained static $\mathcal{M}^{natural}$,
we further generalize our method to Learnable Boundary Guided Adversarial Training (LBGAT) by training $\mathcal{M}^{natural}$ and our required model $\mathcal{M}^{robust}$ at the same time to dynamically adjust the classifier boundary of $\mathcal{M}^{natural}$ and learn the most robustness-friendly one to further help $\mathcal{M}^{robust}$ enhance robustness. To show the flexibility of our method, we incorporate our model into state-of-the-art methods Adversarial Logit Pairing (ALP) \cite{DBLP:journals/corr/abs-1803-06373} and TRADES \cite{zhang2019theoretically} respectively and accomplish remarkable improvement over the baselines. Interestingly, in our exploration, we observe the classifier boundary guidance from $\mathcal{M}^{natural}$ can also enhance model robustness, which gives us new insights and potentially facilitates progress for adversarial robustness. 

We conduct experiments on CIFAR-10, CIFAR-100, and Tiny ImageNet to evaluate the performance of our models under both white- and black-box attacks. Our models achieve impressive performance on these datasets and outperform previous work in a large margin. Particularly, we achieve state-of-the-art model robustness on CIFAR-100 without extra real or synthetic data under current the most popular auto-attack.

\section{Related Work}
\subsection{Adversarial Attack}
\paragraph{White-box Attack}
Szegedy et al.  \cite{DBLP:journals/corr/SzegedyZSBEGF13} observed that CNNs are vulnerable to adversarial examples computed by the proposed box-constrained L-BFGS attack method. Goodfellow et al. \cite{DBLP:journals/corr/GoodfellowSS14} attributed the existence of adversarial examples to the linear nature of networks, which yields the fast gradient sign method (FGSM) for efficiently generating adversarial examples. 

FGSM was further extended to different versions of iterative attack methods. Kurakin et al. \cite{DBLP:conf/iclr/KurakinGB17a} showed that adversarial examples could exist in the physical world with an I-FGSM attack and iteratively applied FGSM multiple times with a small step size.  Madry et al. \cite{DBLP:conf/iclr/MadryMSTV18} proposed Projected Gradient Descent (PGD) method as a universal “first-order adversary”, {\it i.e.}, the most active attack utilizing the local first-order information about the network. 

Dong et al. \cite{DBLP:conf/cvpr/DongLPS0HL18} integrated the momentum term into an iterative process for attack, called MI-FGSM, to stabilize update of directions and escape from poor local maxima during iterations. This method obtains more transferable adversarial examples. Moreover, boundary-based methods like DeepFool \cite{DBLP:conf/cvpr/Moosavi-Dezfooli16} and optimization-based methods like C\&W \cite{DBLP:conf/sp/Carlini017} were also developed, making adversarial defense more challenging. Recently, the ensemble of diverse attack methods -- auto-attack \cite{croce2020reliable} by Croce et al., consisting of APGD-CE \cite{croce2020reliable}, APGD-DLR \cite{croce2020reliable}, FAB \cite{croce2020minimally}, and Square Attack \cite{ACFH2020square}, became popular benchmark for testing model robustness.  

\paragraph{Black-box Attack}
There are also many ways to explore the transferability of adversarial examples for the black-box attack. Liu et al. \cite{DBLP:conf/iclr/LiuCLS17} was the first to study the transferability of targeted adversarial examples. They observed that a large proportion
of target adversarial examples were able to transfer with their target labels using the proposed ensemble-based attack method. Dong et al. \cite{DBLP:conf/cvpr/DongLPS0HL18} showed that iterative attack methods incorporating the momentum term achieved better transferability. Further, Xie et al. \cite{DBLP:conf/cvpr/XieZZBWRY19} boosted the transferability of adversarial examples by creating diverse input patterns with random resize and random padding.

\subsection{Adversarial Defense}
Recent work focuses generally on developing defense methods to improve model robustness, including input transformation-based methods, randomization based methods \cite{DBLP:conf/iclr/XieWZRY18,DBLP:conf/iclr/DhillonALBKKA18}, and adversarial training~\cite{zhang2019theoretically,DBLP:journals/corr/abs-1803-06373,DBLP:conf/iclr/MadryMSTV18,DBLP:conf/iclr/TramerKPGBM18}. 
Athalye et al. \cite{pmlr-v80-athalye18a} showed that adversarial training with PGD had withstood active attacks. Tram{\`{e}}r et al. \cite{DBLP:conf/iclr/TramerKPGBM18} raised model robustness under black-box attack by the proposed ensemble adversarial training, \emph{i.e.}, producing adversarial examples by static ensemble models. 
Madry et al. \cite{DBLP:conf/iclr/MadryMSTV18} used the universal first-order adversary, {\it i.e.}, PGD attack, to obtain adversarial examples in the course of adversarial training. 
Differently, Kannan et al. \cite{DBLP:journals/corr/abs-1803-06373} enhanced model robustness with adversarial logit pairing, which encourages the logits from natural images and adversarial examples to be similar to each other in the same model. 

Moreover, Zhang et al. \cite{zhang2019theoretically} regularized the output from natural images and adversarial examples with the KL-divergence function, meanwhile using a variant of PGD attack. 
Xie et al. \cite{xie2019intriguing} studied the effect of normalization in adversarial training and proposed the Mixture BN mechanism that uses separate batch normalization layers for natural data and adversarial examples in one model. It still requires the strong assumption of knowing whether an image is natural or adversarial, at inference time, which may not be that practical. 

\subsection{Knowledge Distillation}
Knowledge distillation was first used in \cite{DBLP:journals/corr/HintonVD15} by Hinton et al., which was then widely applied to distill knowledge from a teacher model to a student model. The typical application of knowledge distillation is model compression, transferring from a large network or ensembles to a small network that better suits low-cost computing. Since this work, several methods \cite{Tung_2019_ICCV,DBLP:conf/cvpr/ParkKLC19,DBLP:journals/corr/SauB16,DBLP:conf/iclr/TarvainenV17,DBLP:journals/corr/abs-1710-07535, tian2019crd, Chen_2021_CVPR} were proposed to further improve performance on model compressing and other tasks. 

Goldblum et al. \cite{goldblum2019adversarially} analyzed the application of knowledge distillation in adversarial training and proposed Adversarial Robust Distillation (ARD) to transfer robustness from a large adversarially trained model to a smaller one. In this paper, we propose to use one robustness-friendly boundary learned by one natural model, not necessarily large, to guide the adversarial training without cross-entropy loss. By this way, the robust model can sufficiently inherit the classifier boundary and thus preserves high accuracy on natural data.

\section{Our Method}
\subsection{Boundary Guided Adversarial Training}
As suggested by Madry et al. \cite{DBLP:conf/iclr/MadryMSTV18}, projected gradient descent (PGD) is a universal first-order adversary. Robust methods to defense PGD might be able to resist attack stemming from other first-order methods as well. Similarly, we use adversarial training with
PGD as
\begin{equation}
\mathop{\min}_{\theta} \mathbb{E}_{(x,y) \in \hat{p}_{data}} \left( \mathop{\arg \max}_{\delta} \hat{L}(\theta, x+\delta, y) \right)
\end{equation}
where $\hat{p}_{data}$ is the training data distribution,
$\hat{L}(\theta, x, y)$ is the standard cross-entropy loss function with data point $x$ and its corresponding true label $y$. $\theta$ represents parameters of the model, and the maximization with respect to $\delta$ is approximated using noisy BIM~\cite{DBLP:conf/iclr/KurakinGB17a}. We denote the adversarial example $x+\delta$ across the paper as $x^{adv}$. Following previous work~\cite{zhang2019theoretically,DBLP:conf/iclr/MadryMSTV18}, $\delta$ is bounded by $l_{\infty}$.

Our expectation of the robust model is to achieve decent robustness and at the same time keep high accuracy on natural images. 
As illustrated in Fig. \ref{fig:illustration}, we make use of logits from a clean model to help shape the classifier boundary of the robust model. The logits of our required robust model $\mathcal{M}^{robust}$ with $x^{adv}$ taken as input should be similar to those of $\mathcal{M}^{natural}$ taking $x$ as input. This relation is expressed as
\begin{equation}
\mathop{\min}_{\theta} \mathbb{E}_{(x,y) \in \hat{p}_{data}} L\left(\mathcal{M}^{robust}(x^{adv}), \mathcal{M}^{natural}(x)\right) \label{f_BGAT}
\end{equation}
where $L$ is Mean Square Error (MSE) loss function in our experiments and $\mathcal{M}(x)$ denotes the logits of model $\mathcal{M}$ taking $x$ as input. $\theta$ is the parameter of $\mathcal{M}^{robust}$. We randomly initialize $\mathcal{M}^{robust}$ and off-line train $\mathcal{M}^{natural}$ on natural data in our experiments.

Our method can be understood from the perspective of \textit{classifier boundary guidance}. Here we give analysis of why our method can yield high performance on natural data.

\paragraph{Natural Classifier Boundary Guidance} 
Since we assume that $\mathcal{M}^{natural}$ is well trained on natural data, logits from $\mathcal{M}^{natural}$ embed more discriminative features for classification, especially the classifier boundary. According to Eq. \eqref{f_BGAT}, when we impose the logits constraints, the system penalizes more on those pairs ($x$ and $x^{adv}$) that have more substantial discrepancy in classification. Therefore, this logit guidance makes $\mathcal{M}^{robust}$ inherit decent classifier boundary for adversarial data. Actually, the inherited classifier boundary is still applicable to natural data in following explanation.

It is noteworthy that the adversarial example $x^{adv}$ is located in the $l_{\infty}$ ball of $x$. According to the min-max mechanism of PGD~\cite{DBLP:conf/iclr/MadryMSTV18}, when the adversarial training converges, the loss value corresponding to $x^{adv}$ is always larger than the loss value corresponding to $x$ when passing $x^{adv}$ and $x$ into the same model $\mathcal{M}^{robust}$. Therefore, when we pull $x^{adv}$ into the correct class with our proposed logits constraints, $x$ is also squeezed into the correct class. Thus the inherited classifier boundary from $\mathcal{M}^{natural}$ separates natural data well and preserves high natural accuracy. 

\subsection{Learnable Boundary Guided Adv. Training}
\label{sec_LBGAT}
For Boundary Guided Adversarial Training (BGAT) method, $\mathcal{M}^{robust}$ is constrained by logits of the static $\mathcal{M}^{natural}$. The well trained $\mathcal{M}^{natural}$ has the most desirable classifier boundary for natural data. Thus, inheriting such classifier boundary, $\mathcal{M}^{robust}$ tends to achieve high performance on natural images. 

Nevertheless, the classifier boundary coming from static $\mathcal{M}^{natural}$ might not be the most suitable choice for pursuing robustness.
We generalize the BGAT method to Learnable Boundary Guided Adversarial Training (LBGAT) by training $\mathcal{M}^{natural}$ and $\mathcal{M}^{robust}$ simultaneously and collaboratively. The loss function is therefore changed from Eq. \eqref{f_BGAT} to
\begin{small}
	\begin{align}
	\mathop{\min}_{\theta,\theta^{*}} \mathbb{E}_{(x,y) \in \hat{p}_{data}} &L\left(\mathcal{M}^{robust}(x^{adv}), \mathcal{M}^{natural}(x)\right) \nonumber \\
	& + \beta ~ CE\left(\sigma(\mathcal{M}^{natural}(x)),y\right) \label{f_LBGAT}
	\end{align}
\end{small}
where $x^{adv}$ is the adversarial example corresponding to its natural data $x$, and $y$ is the true label. $\sigma(\cdot)$ is a softmax function. CE represents cross-entropy loss, $\mathcal{M}^{natural}$ and $\mathcal{M}^{robust}$ are parameterized by $\theta^{*}$ and $\theta$ respectively. We use Mean Square Error (MSE) loss as L function. $\beta$ is the trade-off parameter. In this paper, we choose $\beta=1$. We randomly initialize $\mathcal{M}^{robust}$ and $\mathcal{M}^{natural}$ in our experiments.

Under the regularization of the proposed logits constraints, {\it i.e.}, the $L(\cdot)$ loss item in Eq.~\eqref{f_LBGAT}, $\mathcal{M}^{natural}$ adaptively learns one most robustness-friendly classifier boundary during the collaborative training. At the same time, it guarantees least performance degradation on natural data with $CE(\cdot)$ loss item in Eq.~\eqref{f_LBGAT}. Note there is no additional cross-entropy loss for optimizing $\mathcal{M}^{robust}$, which makes the classifier boundary be sufficiently inherited from $\mathcal{M}^{natural}$. More details are listed in Algorithm \ref{algorithm_LBGAT}.

\begin{algorithm}[t]
	\caption{Learnable Boundary Guided Adversarial Training (LBGAT)}
	\label{algorithm_LBGAT}
	\begin{algorithmic}[1]
		\State \textbf{Input:} step size $\eta_{1}$ and learning rate $\eta_{2}$, batch size $m$, number of iterations $K$ in inner optimization, model $\mathcal{M}^{robust}$ parameterized by $\theta$, $\mathcal{M}^{natural}$ parameterized by $\theta^{*}$. $\beta$ is one hyper-parameter.
		\State \textbf{Output:} robust model $\mathcal{M}^{robust}$ with $\theta$.
		\State Initialize $\mathcal{M}^{robust}$ and $\mathcal{M}^{natural}$ randomly or with pre-trained configuration.
		\Repeat
		\State Read mini-batch $X=\{x_{1}, ..., x_{m}\}$, $Y=\{y_{1}, ..., y_{m}\}$ from training set;
		\State Get adversarial examples $X^{adv} = \{x_{1}^{adv}, ..., x_{m}^{adv} \}$ by PGD attack with input $X$, $Y$;
		\State $output^{n}=\mathcal{M}^{natural}(X)$;
		\State $output^{r}=\mathcal{M}^{robust}(X^{adv})$;
		\State $loss_{ce}=cross-entropy(\sigma(output^{n}), Y)$;
		\State $loss_{reg}=L(output^{n}, output^{r})$;
		\State $\theta^{*}= \theta^{*} - \eta_{2} \sum_{i=1}^{m} \nabla_{\theta^{*}} (\beta loss_{ce}+ loss_{reg}) /m$;
		\State $\theta= \theta - \eta_{2} \sum_{i=1}^{m} \nabla_{\theta} (\beta loss_{ce}+ loss_{reg}) /m$;
		\Until training converges
	\end{algorithmic}
\end{algorithm}

\subsection{Boundary Guidance Improving Robustness} 
Zhang et al. \cite{zhang2019theoretically} identified a trade-off between performance on natural data and robust accuracy. Xie et al. \cite{xie2020adversarial} observed that adversarial examples were helpful to model generalization ability on natural images. However, using models trained only with natural data to enhance model robustness remains unexplored. We instead notice that proper classifier boundary learned by the naturally trained model not only helps preserve high natural accuracy but also enhances model robustness (\textbf{2.44\%} improvement on CIFAR-100 dataset under the strongest auto-attack \cite{croce2020reliable} shown in Table~\ref{tab:white-box}. We attribute the improvement to the guidance of natural classifier boundary with the following explanation. 

Empirically, as shown in Fig.~\ref{fig:comparison_methods}, an adversarially trained model usually suffer from natural accuracy degradation, which means the adversarially trained model can not model the relations among different classes as well as the naturally trained model.

For example, with an image of a dog, the naturally trained model can misclassify it as a cat with the probability of 0.5. Under some case, we can accept this result because
some dogs are very like a cat in real life. However, the adversarially trained model can misclassify a dog into a truck with high confidence because attackers can change the prediction of an image into any other class. And this is not acceptable for us because a dog is very different from
a truck. Thus, with the guide of classifier boundary from a naturally trained model, the adversarially trained model can avoid such issues to some degree in training optimization.

\subsection{Model Flexibility}
Our method provides a new training scheme for adversarial training. It does not conflict or overlap with other adversarial training methods. We show the flexibility of our approach by using it in other state-of-the-art methods, {\it e.g.}, Adversarial Logit Pairing (ALP)~\cite{DBLP:journals/corr/abs-1803-06373} and TRADES~\cite{zhang2019theoretically}. We validate the improvement over these baselines.

\paragraph{Combined with Adversarial Logit Paring}
Adversarial logit pairing (ALP) requires the logits of natural data $x$ and the corresponding adversarial example $x^{adv}$ to be the same in one model, which is achieved by adding an extra mean square loss item between two logits output. We combine our BGAT with ALP as the loss of
\begin{small}
	\begin{align}
	\mathop{\min}_{\theta} \quad & \mathbb{E}_{(x,y) \in \hat{p}_{data}} L\left(\mathcal{M}^{robust}(x^{adv}), \mathcal{M}^{natural}(x)\right) \nonumber \\
	&+ \alpha~MSE\left(\mathcal{M}^{robust}(x^{adv}), \mathcal{M}^{robust}(x)\right) \label{f_BGAT_ALP}
	\end{align}
\end{small}
where $\alpha$ is a trade-off parameter. $\sigma(\cdot)$ is a softmax function and $y$ is the true label. $\theta$ is the parameter of $\mathcal{M}^{robust}$. We replace the cross-entropy loss item $CE(\sigma(\mathcal{M}^{robust}(x^{adv})),y)$ in the original ALP loss function with our Eq.~(\ref{f_BGAT}).

\paragraph{Combined with TRADES}
The proposed TRADES algorithm~\cite{zhang2019theoretically} explores the trade-off between model robustness and accuracy on natural data by optimizing one regularized surrogate loss. We use our BGAT in the TRADES algorithm as
\begin{small}
	\begin{align}
	\mathop{\min}_{\theta} \quad & \mathbb{E}_{(x,y) \in \hat{p}_{data}} L\left(\mathcal{M}^{robust}(x^{adv}), \mathcal{M}^{natural}(x)\right) \nonumber \\
	&+ \alpha~D_{KL}\left(\sigma(\mathcal{M}^{robust}(x^{adv})) || \sigma(\mathcal{M}^{robust}(x))\right) \label{f_BGAT_TRADES}
	\end{align}
\end{small}
where $\alpha$ is still a trade-off parameter.  $\theta$ is the parameter of $\mathcal{M}^{robust}$. $\sigma(\cdot)$ is softmax function and $y$ is the true label. $D_{KL}(\cdot)$ is the boundary error term, pushing classifier boundary away from data point $x$, originally defined in TRADES~\cite{zhang2019theoretically}. We replace the cross-entropy loss item of  $CE(\sigma(\mathcal{M}^{robust}(x)),y)$ in original TRADES loss with our Eq. (\ref{f_BGAT}). 
%Note that TRADES uses a variant of PGD attack during adversarial training.

It is noted that our LBGAT method can also be combined with both ALP and TRADES methods by simply replacing the first loss item in Eqs. (\ref{f_BGAT_ALP}) and (\ref{f_BGAT_TRADES}) with Eq. (\ref{f_LBGAT}).

\section{Experiments}
\label{exp_evaluation}
In this section, we verify the effectiveness of our methods by conducting both white- and black-box attack following the same experimental settings in \cite{zhang2019theoretically}, {\it i.e.}, applying $FGSM^{k}$ (white-box or black-box) attack with 20 iterations, perturbation size $\epsilon=0.031$  with step size 0.003.

\paragraph{Datasets.}
To evaluate the robustness of our models, we conduct extensive experiments on CIFAR-10, CIFAR-100 and Tiny ImageNet datasets. CIFAR-10 dataset consists of 60,000 32x32 color images in 10 classes, with 6,000 images per class. There are 50,000 training images and 10,000 test images. 
CIFAR-100 has 100 classes containing 600 images each. There are 500 training images and 100 testing images per class. 
Tiny Imagenet \cite{DBLP:conf/cvpr/DengDSLL009}, which is with more complex data, is a miniature of ImageNet dataset. It has 200 classes. Each class has 500 training images, 50 validation images. In our experiments, we resize the image to 32x32 and normalize pixel values to [0,1].
Following \cite{zhang2019theoretically}, we perform standard data augmentation including random crops with 4 pixels of padding and random horizontal flip during training.

\paragraph{Training Details.}
We use the same neural network architecture as \cite{zhang2019theoretically}, {\it i.e.}, the wide residual network WRN-34-10. Following \cite{zhang2019theoretically}, We set perturbation $\epsilon = 0.031$, perturbation step size $\eta_{1} = 0.007$, number of iterations $K = 10$, learning rate $\eta_{2} = 0.1$, batch size $m = 128$, and number of training epochs 100 with transition epochs $\{75,90\}$ on the training dataset. Similarly, SGD optimizer with momentum 0.9 and weight decay $2e-4$ is adopted.

\subsection{Ablation Studies}

\begin{table}
	\centering
	\caption{Ablation study for boundary inheritance on CIFAR-10. 20 iterations PGD white-box attack is applied. We adopt ResNet18 as $\mathcal{M}^{natural}$ for LBGAT method. $Acc_{n}$ represents accuracy on natural images while $Acc_{r}$ represents robustness of models.} 
	\resizebox{0.80\linewidth}{!}
	{
		\begin{tabular}{l|c|c}
			\textbf{Methods} &$\textbf{Acc}_{n}$ &$\textbf{Acc}_{r}$ \\
			\hline
			\hline
			vanilla AT          &86.82\% &52.87\% \\
			TRADES ($\alpha=6$) &84.92\% &56.61\% \\
			LBGAT ($\alpha=0$) (KL)        &88.00\%  &56.10\% \\
			LBGAT ($\alpha=0$) w/          &88.35\% &55.50\% \\
			LBGAT ($\alpha=0$) w/o         &88.22\% &57.55\% \\
			\hline
			\hline
		\end{tabular}
		\label{tab:ablation_boundary_inheritance}
	}
\end{table}

\subsubsection{Natural Classifier Boundary Inheritance}
\label{sec:L_function}
To show the importance of boundary inheritance from $\mathcal{M}^{natural}$, we conduct ablation experiments with and without cross-entropy loss for $\mathcal{M}^{robust}$ in Eq. \eqref{f_LBGAT}. Experimental results are summarized in Table~\ref{tab:ablation_boundary_inheritance}. "w/o" additional cross-entropy loss for $\mathcal{M}^{robust}$ enjoys 2.05\% higher robust accuracy than "w/", which further manifests vast importance of the natural classifier boundary inheritance.  We also replace MSE loss with KL-Divergence loss in Eq. \eqref{f_LBGAT}. KL-Divergence loss encourages the outputs of $\mathcal{M}^{robust}$ and
$\mathcal{M}^{natural}$ to enjoy the same distribution while MSE loss encourages the outputs of $\mathcal{M}^{robust}$ and $\mathcal{M}^{natural}$ to have the same classifier boundary. After replacing MSE with KL-Divergence, we observe performance degradation.

\subsubsection{Feature Visualization}
We randomly sample 5 or 20 classes in CIFAR-100. The numbers in the pictures are class indexes. For each sampled class, we collect the logit features of clean images and the corresponding adversarial examples. As shown in the figures below, LBGAT can inherit a good classifier boundary from a naturally trained model, benefiting performance on both natural data and adversarial data of the adversarially trained model. 

\begin{figure}
	\begin{subfigure}{.23\textwidth}
		\centering
		\includegraphics[width=1.0\linewidth]{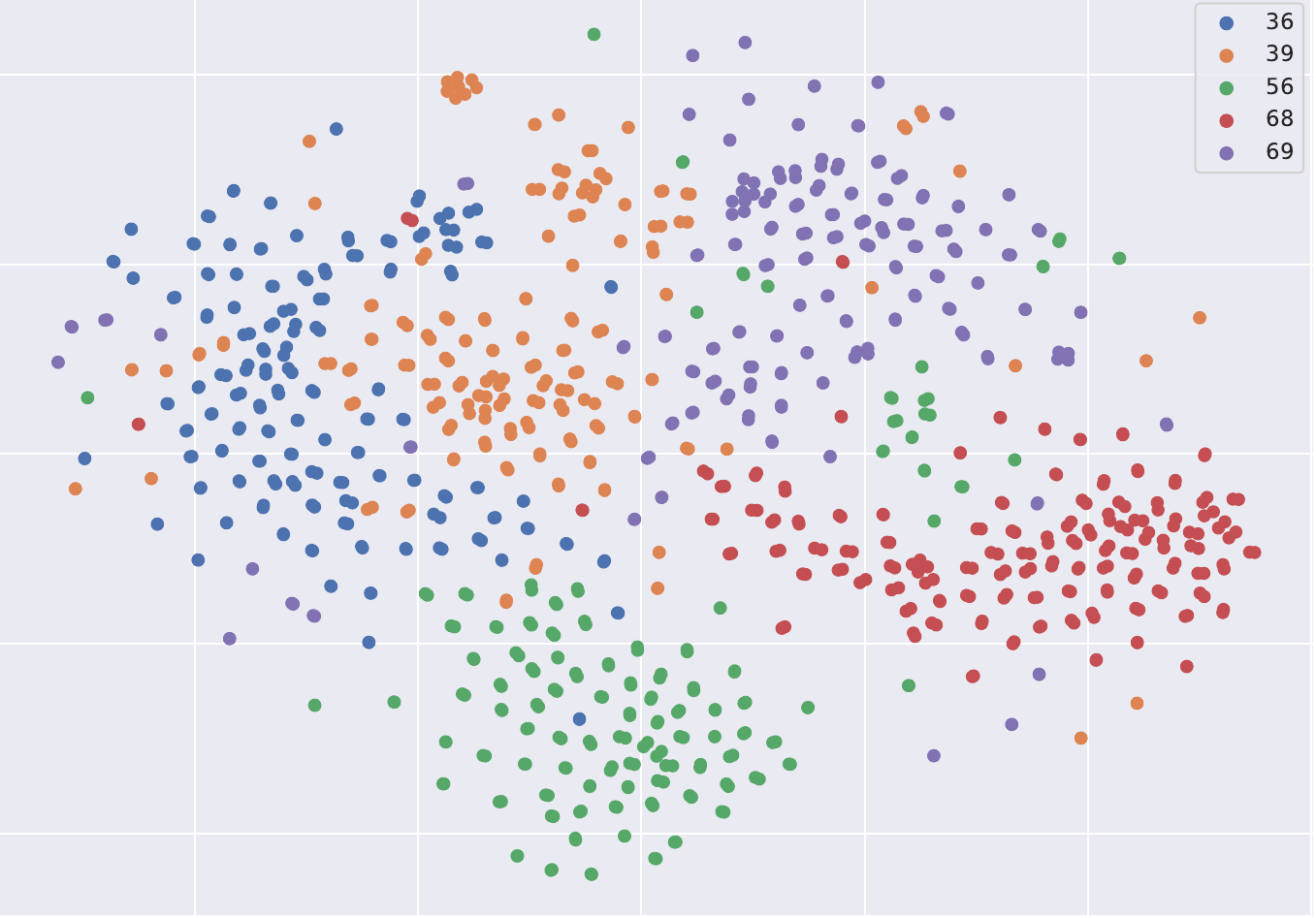}
		\caption{Visualization for TRADES.}
		\label{fig:sub-first}
	\end{subfigure}
	\begin{subfigure}{.23\textwidth}
		\centering
		\includegraphics[width=1.0\linewidth]{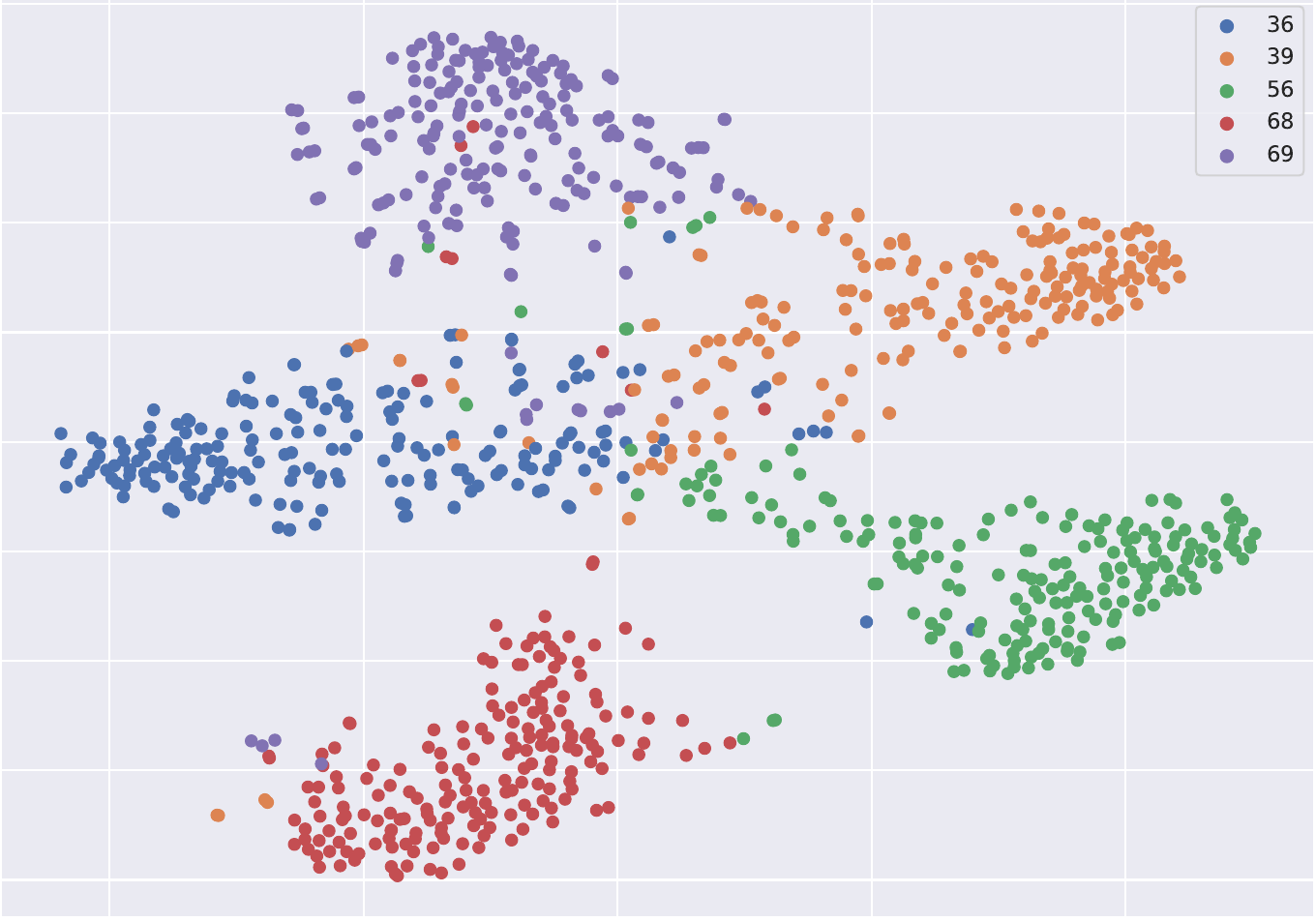}  
		\caption{Visualization for LBGAT.}
		\label{fig:sub-second}
	\end{subfigure}
	\label{fig:tsne_5c}
	\caption{Feature Visualization for LBGAT and TRADES on 5 random selected classes.}
\end{figure}

\begin{figure}
	\begin{subfigure}{.23\textwidth}
		\centering
		\includegraphics[width=1.0\linewidth]{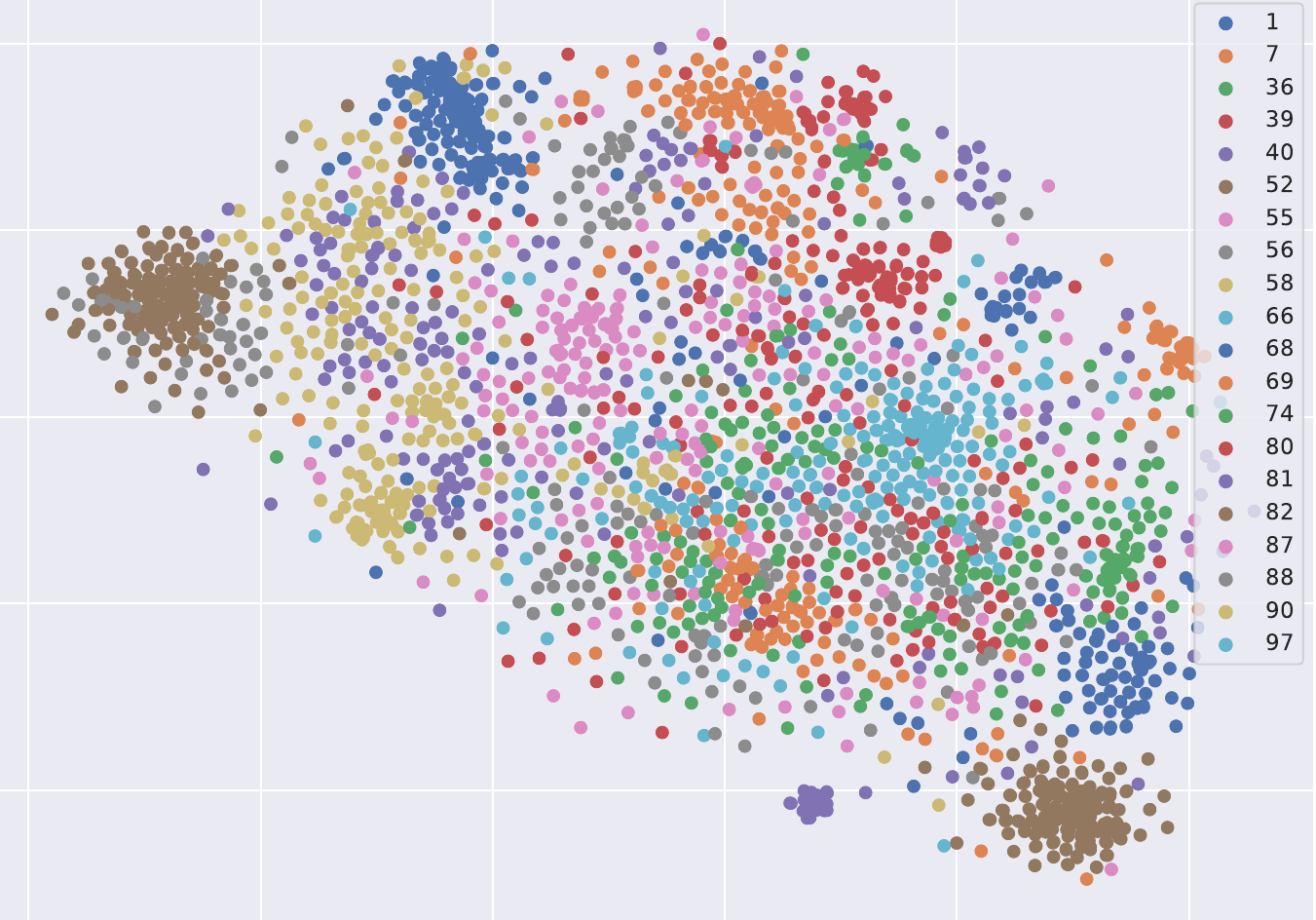}
		\caption{Visualization for TRADES.}
		\label{fig:tsne_trades6_20c}
	\end{subfigure}
	\begin{subfigure}{.23\textwidth}
		\centering
		\includegraphics[width=1.0\linewidth]{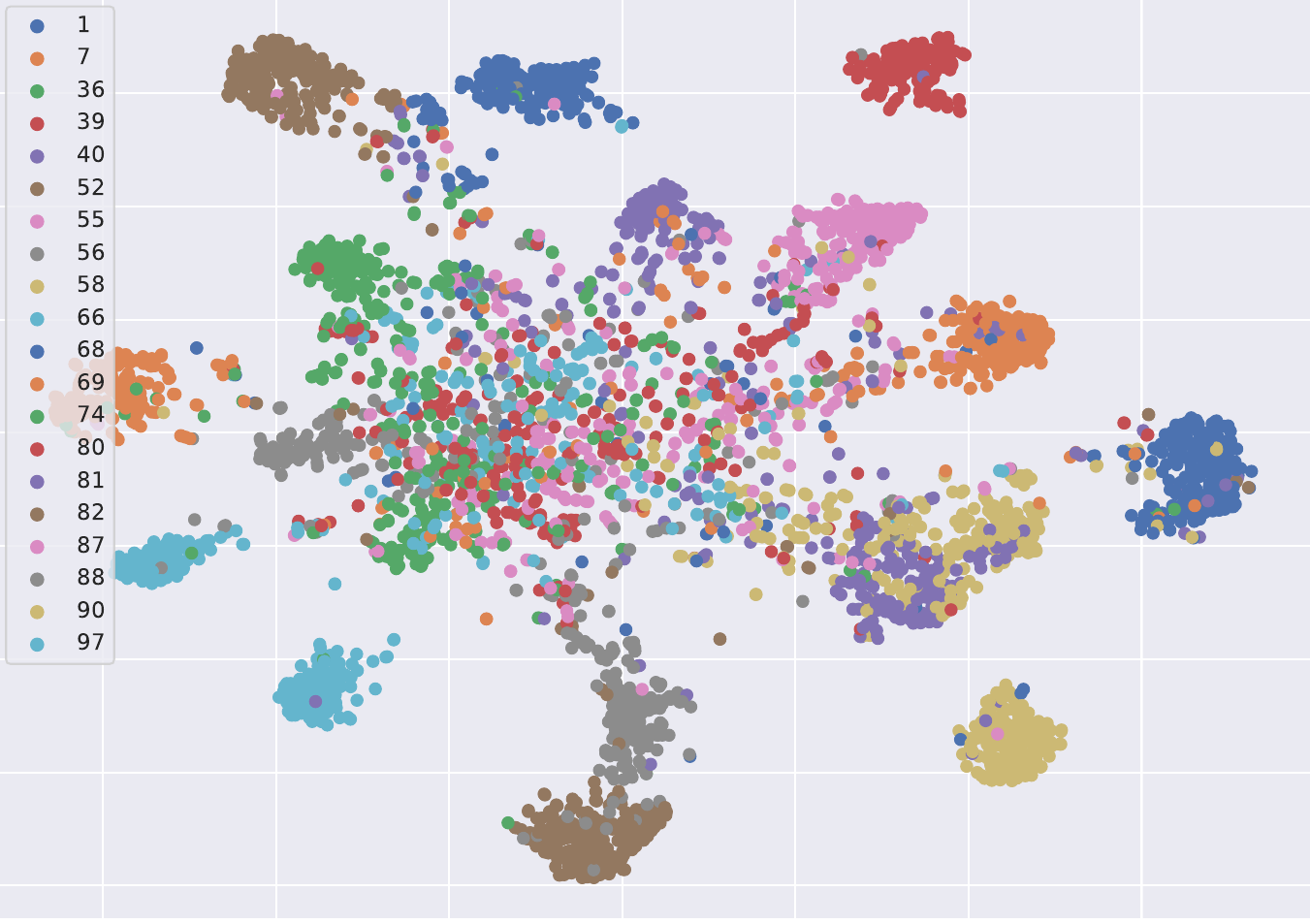}  
		\caption{Visualization for LBGAT.}
		\label{fig:tsne_lbgat0_20c}
	\end{subfigure}
	\label{fig:tsne_20c}
	\caption{Feature Visualization for LBGAT and TRADES on 20 random selected classes.}
\end{figure}

\subsubsection{Separate Batch Normalization}
Xie et al. pointed that clean and adversarial images are drawn from two different domains and disentangling the mixture distribution for normalization can enhance model robustness. However, in this paper, we explore the interaction of information from those two domains, {\textit i.e.}, using classifier boundary information from clean images to assist the learning for adversarial examples.

Here we go deeper to explore whether the convolution weights can be shared in $\mathcal{M}^{natural}$ and $\mathcal{M}^{robust}$ with experiments on CIFAR-100. The experimental results are shown in Table \ref{tab:ablation_separate_bn}. Unfortunately, we observe robustness drops. 

\begin{table}
	\centering
	\caption{Ablation study for separate batch normalization. Robustness is evaluated under auto-attack. \dag denotes models trained with shared convolution and separate batch normalization.} 
	\resizebox{0.90\linewidth}{!}
	{
		\begin{tabular}{l|c|c|c}
			\textbf{Methods} &$\textbf{Acc}_{n}$ &$\textbf{Acc}_{r}$ &\textbf{Datasets}\\
			\hline
			\hline
			LBGAT ($\alpha=0$)        &70.03\%  &27.05\% &CIFAR-100 \\
			LBGAT ($\alpha=6$)        &60.43\%  &29.34\% &CIFAR-100 \\
			\hline
			LBGAT ($\alpha=0$) \dag        &64.89\%  &24.02\% &CIFAR-100 \\
			LBGAT ($\alpha=6$) \dag        &60.62\%  &27.26\% &CIFAR-100 \\
			\hline
		\end{tabular}
		\label{tab:ablation_separate_bn}
	}
\end{table}

\subsubsection{Effectiveness of Our Method}
\label{exp_effectiveness}
We first verify the effectiveness of our method compared with vanilla Adversarial Training (AT). Evaluation of model robustness is under the white-box attack using the same setting as described at the beginning of Sec. \ref{exp_evaluation}. Both our BGAT and LBGAT methods significantly outperform vanilla AT shown by results in Table~\ref{tab:effectiveness}. As analyzed in Sec.~\ref{sec_LBGAT}, the BGAT method can achieve higher natural accuracy while the LBGAT method tends to have stronger robustness. Since we aim to achieve the strongest robustness while preserving natural accuracy as high as possible, we use LBGAT by default.   

\begin{table}[t!]
	\centering
	\caption{Comparison with vanilla AT method. For BGAT, we use the ensemble of WideResNet and InceptionResNetV2 as $\mathcal{M}^{natural}$. ResNet18 as $\mathcal{M}^{natural}$ is for LBGAT on CIFAR-10 and CIFAR-100. $Acc_{n}$ represents accuracy on natural images, while $Acc_{r}$ represents the robustness of models.} 
	\resizebox{0.80\linewidth}{!}
	{
	\begin{tabular}{l|c|c|c}
		\textbf{Methods} &$\textbf{Acc}_{n}$ &$\textbf{Acc}_{r}$ &\textbf{Datasets} \\
		\hline
		\hline
		vanilla AT  &60.90\% &27.46\%  &CIFAR-100 \\
		BGAT        &67.72\% &30.20\%  &CIFAR-100 \\
		LBGAT       &66.29\% &34.30\%  &CIFAR-100 \\
		\hline
		vanilla AT  &86.82\% &52.87\%  &CIFAR-10 \\
		BGAT        &89.00\% &55.40\%  &CIFAR-10 \\
		LBGAT       &87.08\% &56.60\%  &CIFAR-10 \\
		\hline
	\end{tabular}
    }
	\label{tab:effectiveness}
\end{table}

\subsubsection{Combing with ALP and TRADES}
\label{exp_flexibility}
To verify the flexibility of our method, we show that combined with our BGAT and LBGAT methods, ALP and TRADES further improve performance. For ALP, BGAT+ALP and LBGAT+ALP methods, we adopt $\alpha=1$ following the setting in \cite{DBLP:journals/corr/abs-1803-06373}. For the TRADES method, we adopt $\alpha=6$, with which TRADES achieves the best robustness, as demonstrated in \cite{zhang2019theoretically}.

The evaluation is under the white-box attack following the same setting as described at the beginning of Sec. \ref{exp_evaluation}.
We summarize the results in Table ~\ref{tab:flexibility}. Equipped with regularization items of ALP and TRADES, our method can further enhance model robustness. For CIFAR-100, LBGAT+ALP outperforms ALP by 2.92\% and 6.31\% respectively on natural accuracy and robust accuracy under the white-box attack respectively. Meanwhile, the BGAT+TRADES method also outperforms TRADES in terms of both natural accuracy and robustness under the white-box attack for CIFAR-10, which manifests the great flexibility of our method. 

\begin{table}[h]
	\centering
	\caption{Our method is supplementary to ALP and TRADES. For BGAT, we use the ensemble of WideResNet and InceptionResNetV2 model as $\mathcal{M}^{natural}$. ResNet18 is adopted as $\mathcal{M}^{natural}$ for LBGAT+TRADES and LBGAT+ALP. $Acc_{n}$ represents accuracy on natural images while $Acc_{r}$ represents robustness of models.} 
	\resizebox{1.00\linewidth}{!}
	{
	\begin{tabular}{l|c|c|c}
		\textbf{Methods} &$\textbf{Acc}_{n}$ &$\textbf{Acc}_{r}$ &\textbf{Datasets} \\
		\hline
		\hline
		ALP      & 59.75\% & 28.94\% &CIFAR-100\\
		BGAT+ALP  & 63.46\% & 31.27\% &CIFAR-100\\
		LBGAT+ALP & 62.67\% & 35.25\% &CIFAR-100\\
		TRADES ($\alpha=1$)      &62.37\% &25.31\% &CIFAR-100 \\
		TRADES ($\alpha=6$)      &56.51\% &30.94\% &CIFAR-100 \\
		BGAT+TRADES ($\alpha=0$)  &71.27\% &28.70\% &CIFAR-100 \\
		LBGAT+TRADES ($\alpha=0$) &70.03\% &33.01\% &CIFAR-100 \\
		LBGAT+TRADES ($\alpha=6$) &60.43\% &35.50\% &CIFAR-100 \\
		\hline
		ALP       &85.55\% &54.59\% &CIFAR-10\\
		BGAT+ALP  &86.58\% &55.74\% &CIFAR-10\\
		LBGAT+ALP &85.05\% &57.60\% &CIFAR-10\\
		TRADES ($\alpha=1$) &88.64\% &49.14\% &CIFAR-10 \\
		TRADES ($\alpha=6$)       &84.92\% &56.61\% &CIFAR-10 \\
		BGAT+TRADES ($\alpha=0$)  &89.06\% &56.75\% &CIFAR-10 \\
		LBGAT+TRADES ($\alpha=0$) &88.22\% &57.55\% &CIFAR-10 \\
		LBGAT+TRADES ($\alpha=6$) &81.98\% &57.78\% &CIFAR-10 \\
		\hline
	\end{tabular}
    }
	\label{tab:flexibility}
\end{table}

\begin{table*}[t!]
	\centering
	\caption{Comparison of our method with previous defense models under white-box attack on CIFAR-10 and CIFAR-100. We use ResNet18 as $\mathcal{M}^{natural}$ for LBGAT method. $Acc_{n}$ represents accuracy on natural images while $Acc_{r}$ represents robustness of models. AA is the strongest attack, {\it i.e.}, auto-attack \cite{croce2020reliable}. * denotes the model is WRN-34-20.
	}
    {
	\begin{tabular}{l|c|c | c | c |c }
		\hline
		\hline
		\multirow{2}{*}{\textbf{Defense}} & \multirow{2}{*}{\textbf{Attack}} & \multicolumn{2}{c|}{CIFAR-10} & \multicolumn{2}{c}{CIFAR-100} \\
		\cline{3-6}
		& &$\textbf{Acc}_n$ &$\textbf{Acc}_r$ &$\textbf{Acc}_n$ &$\textbf{Acc}_r$ \\
		\hline
		Baseline                      & None  &95.80\% &0\% &78.76\% &0\% \\       
		\hline
		TRADES ($\alpha=1$)           &$FGSM^{20}(PGD)$ &88.64\% & 49.14\% &62.37\% &25.31\% \\
		TRADES ($\alpha=6$)           &$FGSM^{20}(PGD)$ &84.92\% & 56.61\% &56.50\% &30.93\% \\
		LBGAT+ALP                     &$FGSM^{20}(PGD)$ &85.05\% &57.60\% &62.67\% &35.25\% \\
		LBGAT+TRADES ($\alpha=0$)     &$FGSM^{20}(PGD)$ &88.22\% &57.55\% &70.03\% &33.01\%  \\
		LBGAT+TRADES ($\alpha=6$)     &$FGSM^{20}(PGD)$ &81.98\% &57.78\% &60.43\% &35.50\%  \\
		\hline
		TRADES ($\alpha=1$)  &$CW^{20}(PGD)$  &88.64\%  &50.93\% &62.37\% &24.53\% \\
		TRADES ($\alpha=6$)  &$CW^{20}(PGD)$  &84.92\%  &54.98\% &56.50\% &28.43\% \\
		LBGAT+ALP                    &$CW^{20}(PGD)$    &85.05\% &55.78\% &62.67\% &31.97\% \\
		LBGAT+TRADES ($\alpha=0$)    &$CW^{20}(PGD)$    &88.22\% &56.38\% &70.03\% &31.14\% \\
		LBGAT+TRADES ($\alpha=6$)    &$CW^{20}(PGD)$    &81.98\% &55.53\% &60.64\% &31.50\% \\
		\hline
		TRADES ($\alpha=1$)                       &AA &\textbf{88.64\%} &48.11\% &62.37\% &22.24\% \\
		TRADES ($\alpha=6$)                       &AA &84.92\% &52.64\% &56.50\% &26.87\% \\
		LBGAT+TRADES ($\alpha=0$)                 &AA &\textbf{88.22\%} &\textbf{52.86\%} &\textbf{70.03\%} &\textbf{27.05\%} \\
		LBGAT+TRADES ($\alpha=6$)                 &AA &81.98\% &\textbf{53.14\%} &60.43\% &\textbf{29.34\%} \\
		\hline
		LBGAT+TRADES ($\alpha=0$)*                &AA &\textbf{88.70\%} &\textbf{53.58\%} &\textbf{71.00\%} &\textbf{27.66\%} \\
		LBGAT+TRADES ($\alpha=6$)*                &AA &83.61\% &\textbf{54.45\%} &62.55\% &\textbf{30.20\%} \\
		\hline
	\end{tabular}
	\label{tab:white-box}
    }
\vspace{-0.05in}
\end{table*}

\subsection{Robustness on CIFAR-10 and CIFAR-100}
\label{sec:4.1}
\paragraph{White-box Regular Attacks.}
We evaluate the robustness of our models under the white-box attack using the same setting as described at the beginning of Sec. \ref{exp_evaluation}. 
For CIFAR-10, our LBGAT+TRADES ($\alpha=0$) achieves 88.22\% accuracy on natural images, which outperforms TRADES ($\alpha=6$) by 3.3\% at the same time remaining 57.55\% robust accuracy, 0.94\% higher than that of TRADES ($\alpha=6$). 

For CIFAR-100, our LBGAT+TRADES ($\alpha=0$) achieves 70.03\% accuracy on natural images and 33.01\% robust accuracy, improving TRADES ($\alpha=6$) by 13.53\% and 2.08\% respectively. 
Moreover, our LBGAT+TRADES ($\alpha=6$) further boosts robustness to 57.78\% and 35.50\% on CIFAR-10 and CIFAR-100 respectively. 

We also apply several other regular attack methods, like FGSM and CW, to evaluate our models. Compared with TRADES, our proposed methods consistently achieve better accuracy on natural images and stronger robustness on both CIFAR-10 and CIFAR-100 datasets. The details of our results are presented in Table~\ref{tab:white-box}. Note that the CW attack denotes using CW-loss within the PGD framework here. The evaluation under CW attack is also with 20 iterations, step size 0.003 and perturbation $\epsilon=0.031$. 

\paragraph{White-box Auto-Attack (AA).}
\vspace{-0.1in}
Auto-Attack \cite{croce2020reliable} is to reliably evaluate model robustness with an ensemble of diverse strong attack methods, including APGD-CE, APGD-DLR, FAB, and Square Attack. We use the open-source code from \cite{croce2020reliable} to test our models with perturbation size 0.031. The results are listed in Table \ref{tab:white-box}. Compared with TRADES ($\alpha=6$), our LBGAT+TRADES ($\alpha=0$) model improves natural accuracy by \textbf{13.53\%} and \textbf{3.30\%} on CIFAR-100 and CIFAR-10 separately, while achieving comparable robustness.
Our LBGAT+TRADES ($\alpha=6$) model further boosts robust accuracy, obtaining 29.34\% and 53.14\% on CIFAR-100 and CIFAR-10, outperforming TRADES ($\alpha=6$) by \textbf{2.44\%} and \textbf{0.5\%} respectively.

\paragraph{Black-box Attacks.}
We verify the robustness of our models under the black-box attack. We first train models without using adversarial training on the CIFAR-10 and CIFAR-100 datasets. The same network architectures that are specified at the beginning of this section, {\it i.e.}, the WRN-34-10 architecture \cite{DBLP:conf/bmvc/ZagoruykoK16}, are adopted. We denote these models by naturally trained models as (Natural). 

The accuracy of the naturally trained WRN-34-10 model is 95.80\% on the CIFAR-10 dataset and 78.76\% on the CIFAR-100 dataset. We also implement the method proposed in \cite{zhang2019theoretically} on both datasets with their open-source codebase. For both datasets, the $FGSM^{k}$ (black-box) method is applied to attack various defense models. We set $\epsilon = 0.031$ and apply $FGSM^{k}$ (black-box) attack with 20 iterations with step size set to 0.003. Note that the setup is the same as that specified in the white-box attack.

\begin{figure}[t]
	\begin{center}
		\includegraphics[width=0.49\textwidth]{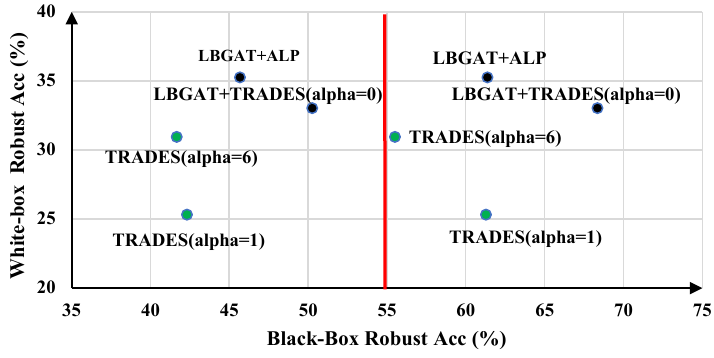}
		\caption{``White-box Robust Acc'' represents classification accuracy under white-box attack. ``Black-box Robust Acc'' represents classification accuracy under black-box attack. Models on the right of the red line are evaluated with the clean model as the source one, while models on the left of the red line models are evaluated with the robust model as the source. More details are included in Table \ref{tab:black-box_cifar100_cifar10}}
		\label{fig:blackbox}
	\end{center}
\vspace{-0.1in}
\end{figure}

The results on CIFAR-100 are summarized in Table ~\ref{tab:black-box_cifar100_cifar10} of Appendix A. We use source models to generate adversarial perturbations where the perturbation directions are according to the gradients of the source models on the input images. Our models are more robust against black-box attack transferred from naturally trained models and TRADES \cite{zhang2019theoretically}, while yielding stronger robustness under white-box attack and higher performance on natural images. Specifically, our best model is \textbf{12.83\%} and \textbf{8.60\%} higher than TRADES ($\alpha=6$) with the naturally trained model and robust model as the source model separately on CIFAR-100. For robustness under black-box attack with one robust source model, our model is tested under TRADES ($\alpha=6$) while TRADES is tested under our LBGAT trained model. More comparison between our method and TRADES is shown in Fig. \ref{fig:blackbox}, which exhibits results on the more challenging dataset CIFAR-100.

\subsection{Robustness on Tiny-ImageNet.}
To further demonstrate the effectiveness of our method on more complex data, we conduct experiments on Tiny ImageNet. Table~\ref{tab:Tiny_ImageNet} shows the experimental results. Our method is better than ALP and TRADES, surpassing baselines with a large margin. Specifically, our LBGAT+TRADES ($\alpha=0$) outperforms the most robust baseline TRADES ($\alpha=6$) by \textbf{9.29\%} on natural data, meanwhile LBGAT+TRADES ($\alpha=6$) is \textbf{3.00\%} higher than it on adversarial data, which verifies the effectiveness of our approach again.

\begin{table}[h]
	\centering
	\caption{Results on Tiny ImageNet~\cite{DBLP:conf/cvpr/DengDSLL009}. The same evaluation setting with CIFAR is applied under 20-iteration PGD white-box attack. We adopt ResNet18 as $\mathcal{M}^{natural}$ for LBGAT methods. $Acc_{n}$ represents accuracy on natural images while $Acc_{r}$ represents robustness of models.} 
	\resizebox{.99\linewidth}{!}
	{
	\begin{tabular}{l|c|c|c}
		\textbf{Methods} &$\textbf{Acc}_{n}$ &$\textbf{Acc}_{r}$ &\textbf{Datasets} \\
		\hline
		\hline
		vanilla AT  &30.65\% &6.81\%  &Tiny ImageNet\\
		LBGAT       &36.50\% &14.00\% &Tiny ImageNet \\
		ALP         &30.51\% &8.01\% &Tiny ImageNet \\
		LBGAT+ALP   &33.67\% &14.55\% &Tiny ImageNet \\
		TRADES ($\alpha=6$)  &38.51\% &13.48\% &Tiny ImageNet \\
		LBGAT+TRADES ($\alpha=0$) &\textbf{47.80\%} &14.31\% &Tiny ImageNet \\
		LBGAT+TRADES ($\alpha=6$) &39.26\% &\textbf{16.42\%} &Tiny ImageNet \\
		\hline
		\hline
	\end{tabular}
	\label{tab:Tiny_ImageNet}
    }
\vspace{-0.1in}
\end{table}

\section{Conclusion}
In this paper, we have proposed the Learnable Boundary Guided Adversarial Training (LBGAT) method, to improve model robustness without losing much accuracy on natural data. Our approach can be understood from the perspective of natural classifier boundary guidance. Moreover, an interesting phenomenon that the boundary guidance from a naturally trained model can also enhance model robustness is observed during our exploration. Finally, extensive experiments on CIFAR-10, CIFAR-100, and more challenging Tiny ImageNet datasets proved the effectiveness of our methods.

{\small
\bibliographystyle{ieee_fullname}
\bibliography{egbib}
}

\newpage
\onecolumn
\appendix

\begin{center}
	\Large \textbf{Learnable Boundary Guided Adversarial Training}
	\Large \\ \textbf{Supplementary Material}
\end{center}
\vspace{20pt}

\vspace{1.0in}	
\section{Robustness under Black-box attack}

\begin{table*}[h]
	\centering
	\caption{Comparison of our method with previous defense models under black-box attack on CIFAR-100 and CIFAR-10. To rule out randomness, the numbers are averaged over 2 independently trained models. $Acc_{n}$ represents accuracy on natural images. $BAcc_{r}$ represents robustness under black-box attack. $WAcc_{r}$ represents robustness under white-box attack}
	\begin{tabular}{l|c|c | c | c|c }
		\textbf{Target Models} &$\textbf{BAcc}_{r}$ &$\textbf{WAcc}_{r}$  & $\textbf{Acc}_{n}$ &\textbf{Source Models}  &\textbf{Dataset}\\
		\hline
		\hline
		TRADES ($\alpha=1$)  &61.29\%  &25.31\%  &62.37\%  &Natural &CIFAR-100\\
		TRADES ($\alpha=6$)  &55.52\%  &30.93\%  &56.51\%  &Natural &CIFAR-100\\
		LBGAT+ALP            &61.38\%  &\textbf{35.25\%}  &62.67\%  &Natural &CIFAR100\\
		LBGAT+TRADES ($\alpha$=0)  &\textbf{68.35\%} &\textbf{33.01\%} &\textbf{70.03\%} &Natural &CIFAR-100\\
		\hline
		TRADES ($\alpha=1$) &42.32\% &25.31\% &62.37\% &LBGAT+TRADES ($\alpha$=0) &CIFAR-100\\
		TRADES ($\alpha=6$) &41.67\% &30.93\% &56.51\% &LBGAT+TRADES ($\alpha$=0) &CIFAR-100\\
		LBGAT+ALP     &45.68\% &\textbf{35.25\%} &62.67\% &TRADES ($\alpha=6$)  &CIFAR-100\\
		LBGAT+TRADES ($\alpha$=0) &\textbf{50.27\%} &\textbf{33.01\%}  &\textbf{70.03\%} &TRADES ($\alpha=6$) &CIFAR-100\\
		\hline
		TRADES ($\alpha=1$)  &87.00\% &49.14\%  &\textbf{88.64\%}  &Natural &CIFAR-10\\
		TRADES ($\alpha=6$)  &83.30\% &56.61\%  &84.92\%  &Natural &CIFAR-10\\
		LBGAT+TRADES ($\alpha=0$)      &\textbf{87.20\%} &\textbf{57.55\%} &\textbf{88.22\%} &Natural &CIFAR-10\\
		\hline
		TRADES ($\alpha=1$) &66.18\% &49.14\% &\textbf{88.64\%} &LBGAT+TRADES($\alpha$=0) &CIFAR-10\\
		TRADES ($\alpha=6$) &67.18\% &56.61\% &84.92\% &LBGAT+TRADES ($\alpha$=0) &CIFAR-10\\
		LBGAT+TRADES ($\alpha=0$)     &\textbf{68.45\%} &\textbf{57.55\%} &\textbf{88.22\%} &TRADES ($\alpha$=6) &CIFAR-10\\
		\hline
		\hline
	\end{tabular}
	\label{tab:black-box_cifar100_cifar10}
\end{table*}

\vspace{0.2in}
\section{Our Method Creates New SOTA Under the Strongest Auto-Attack on CIFAR-100}
To further show the effectiveness of our method, we compare with more previous works. The experimental results are shown in Table~\ref{tab:more_aa_cifar100}. On the more challlenging CIFAR-100 dataset, our method creates a new state-of-the-art (SOTA) on both robustness and natural accuracy. Specifically, our LBGAT ($\alpha=0$) model with WideResNet-34-10 architecture significantly outperforms previous SOAT method \cite{chen2020efficient} by 7.08\% in the aspect of performance on natural data. Meanwhile, our method surpasses it with respect to model robustness. Further, our strongest model LBGAT ($\alpha=6$) with WideResNet-34-10 architecture enjoys 2.4\% higher robustness than \cite{chen2020efficient}.  

Moreover, It is worthy to note that our LBGAT ($\alpha=6$) model achieves even strong robustness than the model, by Hendrycks \etal \cite{hendrycks2019using}, pre-trained on full ImageNet. At the same time, we also surpasses it in the aspect of natural accuracy.   

\vspace{0.2in}
\begin{table*}[h]
	\centering
	\caption{More comparisons under the strongest Auto-Attack on CIFAR-100 dataset. "\dag" denotes numbers are directly copied from \cite{croce2020reliable}. "$\star$" denotes that the method has used additional unlabeled data.} 
	\vspace{0.1cm}
	\begin{tabular}{l|c|c|c}
		\textbf{Methods} & Model &$\textbf{Acc}_{n}$ &$\textbf{Acc}_{r}$ \\
		\hline
		\hline
		LBGAT ($\alpha=0$) Ours                 &WideResNet-34-20 &\textbf{71.00\%} &\textbf{27.66\%} \\
		LBGAT ($\alpha=6$) Ours                 &WideResNet-34-20 &62.55\% &\textbf{30.20\%} \\
		LBGAT ($\alpha=0$) Ours                 &WideResNet-34-10 &\textbf{70.03\%} &\textbf{27.05\%} \\
		LBGAT ($\alpha=6$) Ours                 &WideResNet-34-10 &60.43\% &29.34\% \\
		\hline
		\hline
		
		TRADES ($\alpha=1$) \cite{zhang2019theoretically}      &WideResNet-34-10 &62.37\% &22.24\% \\
		TRADES ($\alpha=6$) \cite{zhang2019theoretically}      &WideResNet-34-10 &56.50\% &26.87\% \\
		Sitawarin \etal \cite{sitawarin2020improving} \dag &WideResNet-34-10 &62.82\%    &24.57\% \\
		Chen \etal \cite{chen2020efficient} \dag           &WideResNet-34-10 &62.15\%	&26.94\% \\ 
		Hendrycks \etal \cite{hendrycks2019using} \dag $\star$     &WideResNet-28-10 &59.23\%	&28.42\% \\
		Rice \etal \cite{rice2020overfitting} \dag         &ResNet-18        &53.83\%    &18.95\% \\
		\hline
		\hline
	\end{tabular}
	\label{tab:more_aa_cifar100}
\end{table*}

\newpage
\section{More Comparisons Under the Strongest Auto-Attack on CIFAR-10}
We also compare with more previous methods on CIFAR-10 dataset. The experimental results are summarized in Table~\ref{tab:more_aa_cifar10}. Our LBGAT ($\alpha=0$) model with WideResNet-34-10 architecture can consistently enjoy higher natural performance while keeping the strongest robustness. We observe that though many fast adversarial training methods, like \cite{wang2019bilateral,shafahi2019adversarial} are proposed to accelerate the training process, their performance are usually unsatisfied.

\begin{table*}[h]
	\centering
	\caption{More comparisons under the strongest Auto-Attack on CIFAR-10 dataset.  "\dag" denotes numbers are directly copied from \cite{croce2020reliable}. "$\ast$" denotes the methods aiming to accelerate adversarial training.} 
	\vspace{0.1cm}
	\begin{tabular}{l|c|c|c}
		\textbf{Methods} & Model &$\textbf{Acc}_{n}$ &$\textbf{Acc}_{r}$ \\
		\hline
		\hline
		LBGAT ($\alpha=0$) Ours                 &WideResNet-34-20 &\textbf{88.70\%} &\textbf{53.58\%} \\
		LBGAT ($\alpha=6$) Ours                 &WideResNet-34-20 &83.61\% &\textbf{54.45\%} \\
		LBGAT ($\alpha=0$) Ours                 &WideResNet-34-10 &\textbf{88.22\%} &\textbf{52.86\%} \\
		LBGAT ($\alpha=6$) Ours                 &WideResNet-34-10 &81.98\% &53.14\% \\
		\hline
		\hline
		Rice \etal \cite{rice2020overfitting} \dag    &WideResNet-34-20 &85.34\%	&53.42\% \\
		TRADES ($\alpha=1$)                            &WideResNet-34-10 &\textbf{88.64\%} &48.11\%\\
		TRADES ($\alpha=6$)                            &WideResNet-34-10 &84.92\%    &52.64\% \\ 
		Kumari \etal \cite{kumari2019harnessing} \dag &WideResNet-34-10 &87.80\%	&49.12\% \\
		Mao \etal \cite{mao2019metric} \dag           &WideResNet-34-10 &86.21\%	&47.41\% \\
		Zhang \etal \cite{zhang2019you} \dag $\ast$   &WideResNet-34-10 &87.20\%	&44.83\% \\
		Shafahi \etal \cite{shafahi2019adversarial} \dag $\ast$ &WideResNet-34-10   &86.11\% &41.47\% \\
		Chan \etal \cite{chan2019jacobian} \dag                 &WideResNet-34-10 &\textbf{93.79\%} &0.26\% \\
		Wang \etal \cite{wang2019bilateral} \dag $\ast$         &WideResNet-28-10 &\textbf{92.80\%} &29.35\% \\
		Qin \etal \cite{qin2019adversarial} \dag         &WideResNet-40-8  &86.28\% &52.81\% \\
		Chen \etal \cite{chen2020adversarial} \dag       &ResNet-50        &86.04\% &51.56\% \\
		Xiao \etal \cite{xiao2019enhancing} \dag         &DenseNet-121     &79.28\% &18.50\% \\
		Wong \etal \cite{wong2020fast} \dag              &ResNet-18        &83.34\% &43.21\% \\
		\hline
		\hline
	\end{tabular}
	\label{tab:more_aa_cifar10}
\end{table*}
\newpage

\end{document}